\documentclass[12pt,draftclsnofoot,onecolumn]{IEEEtran}
\IEEEoverridecommandlockouts 

\usepackage[utf8]{inputenc}
\usepackage{mathptmx} 
\usepackage{amsmath,amssymb} 
\usepackage{dsfont}
\usepackage{verbatim}
\usepackage{multicol,longtable}
\usepackage{epsfig}
\usepackage{txfonts}
\usepackage{calc}
\usepackage{enumerate}
\usepackage{bm}
\usepackage{color}
\usepackage{url}
\usepackage{algorithm}
\usepackage[noend]{algpseudocode}
\usepackage[table]{xcolor}

\newcommand{\cbra}[1]{\ensuremath{\left\{ #1\right\}}}
\newcommand{\abra}[1]{\ensuremath{\left< #1\right>}}
\newcommand{\pder}[2]{\ensuremath{\frac{\partial #1}{\partial #2}}}

\begin{document}

\title{Interpretable machine learning models: a physics-based view}


\author{ Ion Matei, Johan de Kleer, Christoforos Somarakis, Rahul Rai and John S. Baras
}

\maketitle
\begin{abstract}
To understand changes in physical systems and facilitate decisions, explaining how model predictions are made is crucial. We use model-based interpretability, where models of physical systems are constructed by composing basic constructs that explain locally how energy is exchanged and transformed. We use the port Hamiltonian (p-H) formalism to describe the basic constructs that contain physically interpretable processes commonly found in the behavior of physical systems. We describe how we can build models out of the p-H constructs and how we can train them. In addition we show how we can impose physical properties such as dissipativity that ensure numerical stability of the training process. We give examples on how to build and train models for describing the behavior of two physical systems: the inverted pendulum and swarm dynamics.
\end{abstract}

\section{Introduction}

The necessity
for interpretability comes from the fact that it is not always enough to train and model and get an answer, but is also important to understand why a particular answer was given. A simple but meaningful definition of model interpretability given in \cite{miller2017explanation} relates this notion to \textit{the degree to which a human can understand the cause of a decision}. In our case, since we care about models that describe the behavior of physical systems, we change the definition to \textit{the degree to which a human can understand the physical processes that cause  a prediction}.  Throughout this paper we focus on \textit{physically-interpretable} models: models that embed physical laws that explain how energy is transformed and exchanged in the system. A physically-interpretable model facilitates learning and updating the model when something unexpected happens. This update is done by finding an explanation for an unexpected event. For
example, an electrical motor unexpectedly overheats and we ask ourselves: ``Why is the motor overheating?''. We learn that the motor overheats
every time every we subject it to a load above some threshold. Consequently, we can update the model and decide that the motor should not be subjected to a high load or that we should use a different motor rated for higher loads. Such physical interpretability is important for any machine learning (ML) model giving predictions without explanations, where scientific findings stay completely
hidden. Explainable predictions are crucial to facilitate failure analysis, failure progress or  design feedback.

In diagnosis applications \cite{deKleer92,DEKLEER200325,Patton_2000} we would like to explain/solve inconsistencies between our knowledge induced expectations and the observed behavior. For example, we may find out that  there is a contradiction between the knowledge about the vehicles’s past behaviour and the new observations about the current vehicle mileage. Consequently we may ask: ``Why does my vehicle suddenly makes worse mileage, even though it has never done so before?''. The  explanation of the mechanics helps the vehicle's owner reconcile the contradiction: ``One of the brake pads was stuck and consequently the engine had to generate more torque to cope with the additional friction-induced load.''
The more an algorithm/model prediction affects the physical world, the more important is  for the
algorithm/machine to explain its behaviour.

ML models such as classifiers are becoming more ubiquitous in fault detection for physical systems. Take for instance the fault detection and isolation for a wind-turbine. We would like the prediction model, e.g., the classifier, to predict faults with 100\% accuracy, since failing to predict failures can lead to catastrophic events. An explanation  might reveal that the most important feature learned for predicting generator bearing failures is the generator's electrical current at high frequencies, which is indicative of the presence of vibrations in the generator shaft due to bearing (incipient) failures. ML models tend to pick up biases from the training data. Such a phenomenon can turn the
ML models (e.g., classifiers) to favor more common faults, discriminating against rare but possibly catastrophic faults. Interpretability
is a useful debugging tool to detect bias in ML models. Interpretability enables changes in the loss function (e.g., adding new regularization terms) that capture biases in the training data that otherwise would be ignored by the loss function,
which the ML model optimises. An additional reason for demanding interpretability from models is to ensure adoption by industry. Even today, classifiers based on decision trees are popular due to their ability to explain how predictions are made.

Interpretability is one of the main traits behind fault diagnosis and prognostics. Having an interpretation for a faulty prediction helps with the understanding of
the cause of the fault. In addition, it gives an avenue for repairing the system. Interpretability can also be used for diagnosing the ML model itself. From this perspective, we would like to understand why some predictions were incorrect (e.g., misclassification), and fix the model to improve the prediction accuracy. For example, this avenue can be used to generative adversarial networks more  robust. The repair process may include adding/removing features (e.g., sensor measurement) that enables a better discrimination between classes.

We do not always need model interpretation. Such reasons may include: not having a significant impact, the problem is well understood, or interpretability may enable ``gaming'' the system \cite{molnar2019}. An example of the first reason is feedback control design. For such an application a black-box, regression type of model is typically sufficient. The last reason has a significant impact on cyber-physical system security, where an attacker may use the model to understand the physical system and design attack schemes that leverage system weaknesses.

The way we look at the notion of interpretbility in this paper fits also with the view of  \cite{Murdoch_2019}, where interpretability translates to the extraction of insights for a particular audience into a chosen domain problem about domain relationships contained in data. In particular, the insights we produce will be in  mathematical equations format, with physical meaning. Our interpretability method is not  based on post hoc interpretations of deep learning models \cite{8397411,guidotti2018survey}. Our models are not typical regression or statistical models, but they  show how the energy of a system is transformed and exchanged. Namely, they are compositions of basic constructs that describe locally how energy is transformed as it passes through the system.

Among the many criteria used to classify interpretability (see for instance page 16 of \cite{molnar2019}), our models are global, model-based, model specific, and intrinsically interpretable. Model-based interpretability is based on imposing constraints on the form of the ML models so that they  provide meaningful information about the learned relationships. In ML applications there is often a trade-off between the choice of a simpler but easier to interpret model and a more complex (e.g., black box) model, but with low interpretability. The models we propose can be arbitrarily complex but they will still retain the physical interpretability.

We achieve physical interpretability by using a well defined mathematical formalism called the port-Hamilonian (p-H) formalism \cite{h18,h1,h47}. This is a general and powerful geometric framework to model complex dynamical networked systems. P-H systems are based on an energy function (Hamiltonian) and on the interconnection of atomic structure elements (e.g. inertias, springs and dampers for mechanical systems) that interact by exchanging energy. Such models give insights into the physical properties of the system, the framework being particularly suited for finding symmetries (e.g., discrete, or Lie groups of transformations) and conservation laws (under the form of Casimir functions). The models we learn are non-causal in nature since they deal with energy exchanges and transformations and not with changes in the outputs as a result of varying the inputs.


\textbf{Paper Structure:} In Section \ref{interpretability_constructs} we describe the main constructs used to build physically interpretable models. In Section  \ref{interpretable_machine_learning_models} we demonstrate how we can use these constructs to build models for predicting physical system behaviors. We discuss aspects of training p-H based and stable models in Section \ref{training_interpretable_models}. In Section \ref{larger_context} we discussed how our approach fits the broader category of ML interpretable models. We end the paper with two modeling and learning examples (the inverted pendulum and the swarm dynamics) in Section \ref{examples} and some conclusions.

\section{Interpretability constructs}
\label{interpretability_constructs}
In this section we briefly introduce the p-H formalism, its constructs and give some examples of simple physical systems represented in the p-H formalism.

\subsection{Port-Hamiltonian framework}
Consider a finite-dimensional linear state space $\mathcal{X}$
along with a Hamiltonian
$H:\mathcal{X}\rightarrow\mathbb{R}_+$
defining energy-storage,
and a set of pairs of \emph{effort} and \emph{flow} variables
$\cbra{(e_i,f_i)\in\mathcal{E}_i \times \mathcal{F}_i,~i\in\cbra{S,R,P}}$,
describing ports (ensembles of elements) that interact by exchanging energy. The letters ``S'', ``R'' and ``P'' refer to energy storing, resistive and external ports, respectively.
Then, the dynamics of a p-H system
$\Sigma = (\mathcal{X}, H, \mathcal{S}, \mathcal{R}, \mathcal{P},\mathcal{D})$ 
are defined by a Dirac Structure $\mathcal{D}$ \cite{h1,h2} as
$$(f_\mathcal{S},e_\mathcal{S},f_\mathcal{R},e_\mathcal{R},f_\mathcal{P},e_\mathcal{P}) \in \mathcal{D}\Leftrightarrow
e_\mathcal{S}^T f_\mathcal{S} + e_\mathcal{R}^T f_\mathcal{R} +
e_\mathcal{P}^T f_\mathcal{P} = 0,$$
where (\textit{i}) $\mathcal{S} = (f_\mathcal{S},e_\mathcal{S})\in
\mathcal{F}_R \times \mathcal{E}_R =
\mathcal{X} \times \mathcal{X}$
is an energy-storing port, consisting of the union of all the energy-storing elements of the system
(e.g. inertias and springs in mechanical systems),
satisfying $f_\mathcal{S} = -\dot x, e_\mathcal{S} = \pder{H}{x}(x),~x\in\mathcal{X}$ such that
$\frac{d}{dt}H = -e_\mathcal{S}^T f_\mathcal{S}
= e_\mathcal{R}^T f_\mathcal{R} + e_\mathcal{P}^T f_\mathcal{P}$, (\textit{ii}) $\mathcal{R} = (f_\mathcal{R},e_\mathcal{R})\in \mathcal{F}_R \times \mathcal{E}_R$
is an energy-dissipation (resistive) port, consisting of the union of all the resistive elements
of the system (e.g. dampers in mechanical systems), satisfying
$\abra{e_\mathcal{R},f_\mathcal{R}} \leq 0$ and, usually, an input-output relation
$f_\mathcal{R} = -R(e_\mathcal{R})$, (\textit{iii}) $\mathcal{P} = (f_\mathcal{P},e_\mathcal{P})\in \mathcal{F}_P \times \mathcal{E}_P$
is an external port modeling the interaction of the system with the environment,
consisting of a control port $\mathcal{C}$ and an interconnection port $\mathcal{I}$, and (\textit{iv}) $\mathcal{D}\subset \mathcal{F} \times \mathcal{E} =
\mathcal{F}_R \times \mathcal{E}_R \times \mathcal{F}_R \times \mathcal{E}_R\times \mathcal{F}_P \times \mathcal{E}_P$
is a central power-conserving interconnection
(energy-routing) structure (e.g. transformers in electrical systems), satisfying
$ \abra{e,f} =~0,~\forall (f,e)\in\mathcal{D} $, and $ dim\mathcal{D} = dim\mathcal{F}$,
where $\mathcal{E} = \mathcal{F}^*$, and the duality product $\abra{e,f}$ represents power.

The basic property of p-H systems is that the power-conserving interconnection of any number of p-H systems is again a p-H system.  An important and useful special case is the class of input-state-output p-H systems $\dot{x} = [J(x)-R(x)] \frac{\partial H}{\partial x}  + g(x)u$, $y = g^T(x) \frac{\partial H}{\partial x} (x)$,
where $u,y$ are the input–output pairs corresponding to the control port $\mathcal{C}$,
$J(x) = - J^T(x)$ is skew-symmetric, while the matrix $R(x) = R^T (x) \geq 0$
specifies the resistive structure.
%

\subsection{Constructs}

\begin{figure}[ht!]
\begin{center}
\includegraphics[width=0.7\textwidth]{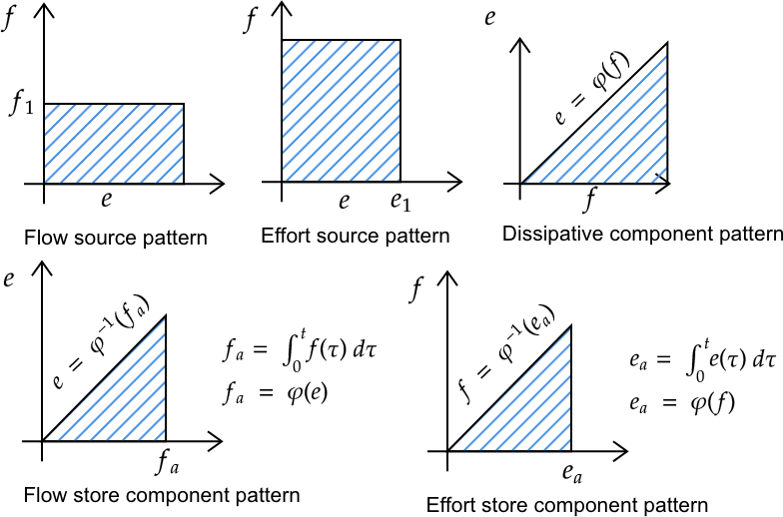}
\end{center}
\caption{Port-Hamiltonian types: flow source construct, effort source construct, resistive (dissipator) construct, flow store construct, effort store construct.}
\label{fig:12191907}
\end{figure}

The Dirac structure operator that enforces the conservation of energy involves a set of constructs/elements with particular ways of transforming energy: energy-storing elements, resistive elements, source elements. With the exception of resistive elements, the energy-storing and source constructs can be defined for both flow and effort type of variables. Figure \ref{fig:12191907} shows examples of patterns of dependencies between the flow and effort variables. Here we are interested mainly in the energy storing and resistive constructs.
Alternative mathematical definitions for flow store, effort store and resitive constructs are shown in equations (\ref{eq:11381222}), (\ref{eq:11391222}) and (\ref{eq:11401222}), respectively.
\begin{equation}
\label{eq:11381222}
\left\{
\begin{array}{ccc}
    \dot{x} &=& f,  \\
     e &=& \frac{\partial H}{\partial x},
\end{array}
\right.
\end{equation}

\begin{equation}
\label{eq:11391222}
\left\{
\begin{array}{ccc}
    \dot{x} &=& e,  \\
     f &=& \frac{\partial H}{\partial x},
\end{array}
\right.
\end{equation}

\begin{equation}
\label{eq:11401222}
R(e,f) = 0, \ p = e^T f\geq 0,
\end{equation}
where $e$ and $f$ are the effort and flow variables, respectively, $H(x)$ is the Hamiltonian function, and $R$ is a resistive function such that (by convention) the instantaneous power $p=e^Tf$ is positive.
The previous construct types are one-port or two-terminal devices. They are not always enough and hence we also use  two-port devices: transformers and gyrators. The constitutive relations for the transformers and gyrators are show in equations (\ref{eq:12281222})-(\ref{eq:12291222}), respectively.
\begin{equation}
\label{eq:12281222}
\left\{
\begin{array}{ccc}
    e_1 &=& g_1(e_2),  \\
     f_1 &=& g_2(f_2)
\end{array}
\right.
\end{equation}
\begin{equation}
\label{eq:12291222}
\left\{
\begin{array}{ccc}
    e_1 &=& g_1(f_2),  \\
     f_1 &=& g_2(e_2)
\end{array}
\right.
\end{equation}
This type of construct classification is domain agnostic, that is, it can be applied to multiples physical domains such as mechanical (translational and rotational), thermal, fluid and magnetics domains.

\subsection{Examples}
In this section we present two examples of physical systems from the mechanical and electrical domain represented using the constructs described in the previous section. We consider the mass-spring damper example with a force as input acting on the mass and the RLC serial circuit.
\subsubsection{Port-Hamiltonian representation of the mass-spring-damper system}
\begin{figure}[ht!]
\begin{center}
\includegraphics[width=0.6\textwidth]{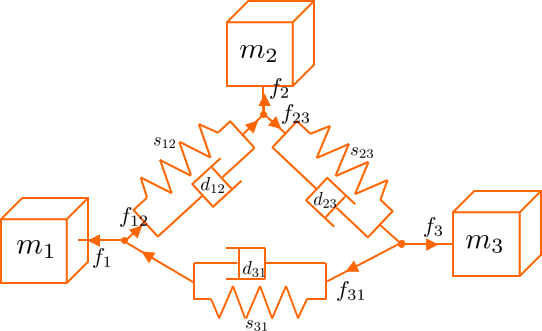}
\end{center}
\caption{Mass-spring-damper system}
\label{fig:12241309}
\end{figure}

The mass-spring damper physical system is depicted in Figure \ref{fig:12241309}. It consists of three components and a force source.

The damper is a resistive construct, whose constitutive equation is $f_d = d\cdot e_d$, where $d$ is the damping constant, $f_d$ is the flow variable (the force), and $e_d$ is the effort variable (the velocity). The spring's Hamiltonian is $H(q) = \frac{1}{2}kq^2$, where $q$ is the spring elongation and $k$ is the stiffness constant. Similar to the damper, case the flow variable is the force $f_k = \frac{\partial H}{\partial q} = kq$, and the effort variable is the relative velocity $e_k = \dot{q}$. Hence, the spring is an effort storage construct. The mass's Hamiltonian in terms of the momentum $p$ is $H(p) = \frac{1}{2m}p^2$, where $m$ is the mass constant.
The mass velocity  acting on the mass is expressed as $e_m =\frac{\partial H}{\partial p} = \frac{1}{m}p$, and the force $f_m=\dot{p}$. Hence it follows that the mass is a flow storage construct. The system has a flow source, as well, defined by the force $f_s = F(t)$. The Dirac structure conserves the system's energy and, assuming that the damper and the spring are grounded at the zero position, is given by $f_m+f_k+f_d = f_s$ and $e_m=e_k=e_d=e_s$. It can be easily check that these effort and flow constraints do satisfy the definition of a Dirac structure.

\subsubsection{Port-Hamiltonian representation of the RLC circuit}
\begin{figure}[ht!]
\begin{center}
\includegraphics[width=0.5\textwidth]{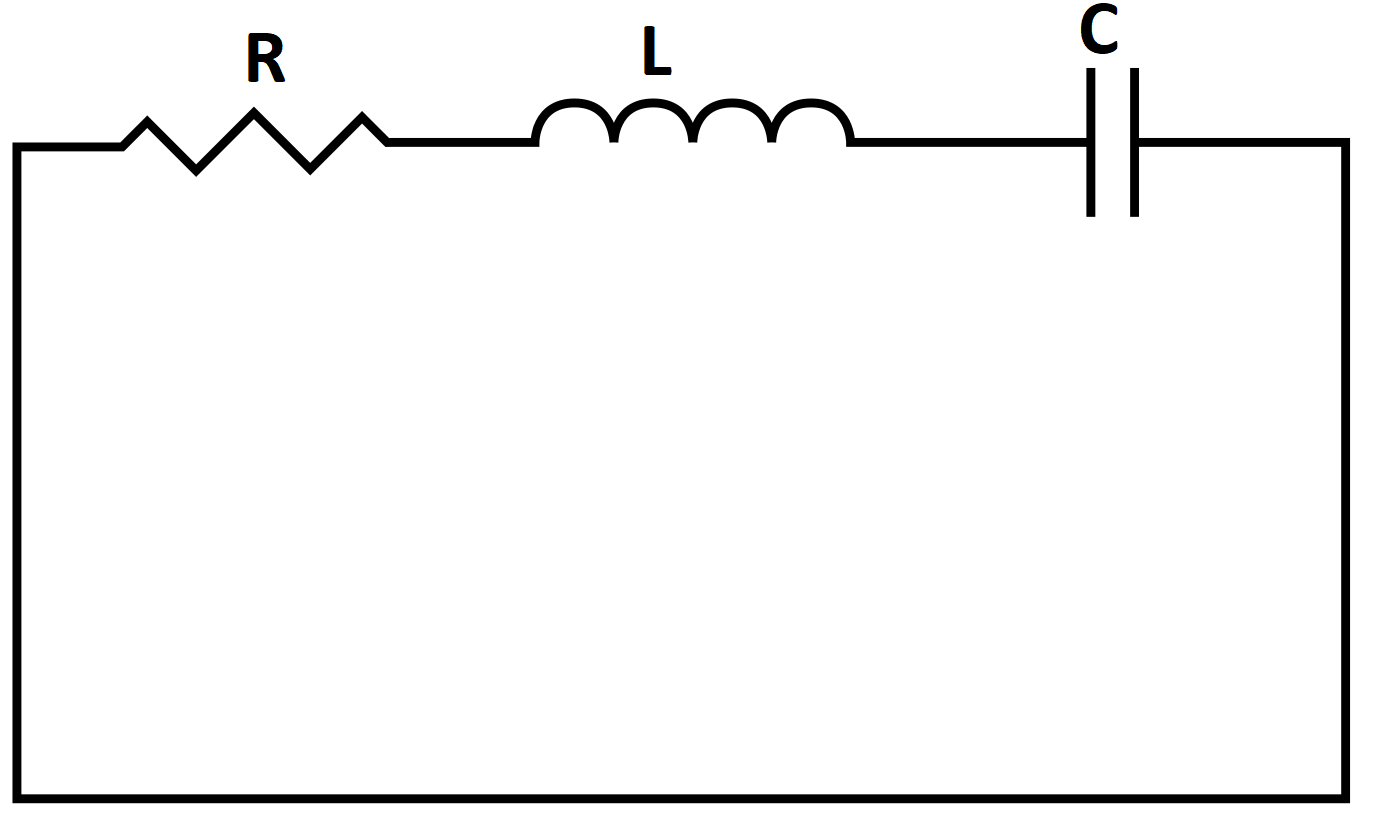}
\end{center}
\caption{RLC circuit}
\label{fig:12271808}
\end{figure}

Consider the RLC circuit in Figure \ref{fig:12271808}. In this example the flow and effort variables are the current and the voltage, respectively.
The resistor constitutive equation is $e_R = R\cdot f_R$, hence it is a resistive construct. The capacitor energy is given by $H(q) = \frac{1}{2C}q^2$, where $C$ is the capacitance, and $q$ is the electric charge. Hence $e_C = \frac{\partial H}{\partial q} = \frac{q}{C}$ and $f_C = \dot{q}$, which shows that the capacitor is a flow storage construct. The inductor's energy is given by $H(\phi) = \frac{1}{2L}\phi^2$, where $L$ is the inductance and $\phi$ is the magnetic flux. The flow is given by $f_L = \frac{\partial H}{\partial \phi} = \frac{\phi}{L}$, and the effort is $e_L = \dot{\phi}$. Therefore, the inductor is an effort storage element. The Dirac structure is given by $f_C=f_R=f_L$ and $e_C+e_L+e_R = 0$, which results in the energy conservation.

\section{Interpretable Machine Learning Models}
\label{interpretable_machine_learning_models}
\begin{figure}[ht!]
\begin{center}
\includegraphics[width=0.5\textwidth]{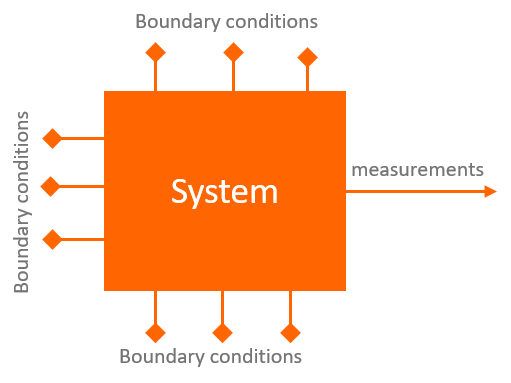}
\end{center}
\caption{Physical system with boundary conditions ports and measurements}
\label{fig:12281157}
\end{figure}

\begin{figure}[ht!]
\begin{center}
\includegraphics[width=0.4\textwidth]{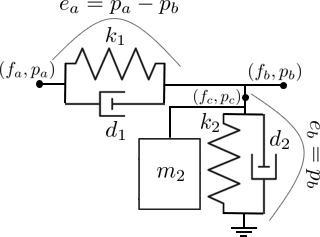}
\end{center}
\caption{Basic construct for building a network of port-Hamiltonian construct expressed in terms of mechanical components.}
\label{fig:12281208}
\end{figure}

The type of systems we can model using the p-H formalism are characterized by boundary conditions through which energy is transferred to the system (e.g., through flow or effort sources) and measurements of system variables (Figure \ref{fig:12281157}). The measurements provide knowledge about the system behavior. The internal behavior of the system is modeled as compositions of p-H constructs based on some given or assumed topology. In the case we know the structural decomposition of the system, we can use this information to construct a topology. When such information is not available, we choose a rich enough topology that has a good chance to capture the observed behavior
Similar to neural networks, we can define a basic construct out of which we can build a network of p-H constructs. In the NN case the basic construct is the linear layer followed by a nonlinear map. Here we define the basic construct in terms of how the flow and the effort variables are manipulated. One possible example of such a construct, expressed in terms of mechanical components, is shown in Figure \ref{fig:12281208}. In this design, we have a resistive (damper $d_1$) and an effort storage element (spring $k_1$) in parallel connected to a mass-spring-damper ($m_2$, $k_2$, $d_2$), where the damper and the spring are grounded.
This design serves two purposes: (i) to allow for a change in the effort variable (the relative velocity between the two ports), and (ii) to allow for a change in the flow variable through the grounded elements. Each component is characterized by a parametrized map: the resistive map for the damper ($R(e,f;w)$), and the energy functions for the storage elements ($H(x;w)$), where $w$ is a vector of parameters. In the case we choose to defined the resistive and energy maps as NN, the vector of parameters $w$ includes the weights and the biases of the NN. In this way, we can define nonlinear masses, springs and dampers. The Dirac structure corresponding to this construct is $f_a+f_b+f_c=0$, and $p_b=p_c$.  To ensure that the construct is dissipative, we enforce (by convention) a positive instantaneous power for the resistive element, that is, $p = e^Tf\geq 0$.  In the case if linear masses-springs-dampers, where the damping coefficient is non-negative,  such a condition is always satisfied. The dissipative conditions can be included as regularization functions or can be directly enforced as inequality constraints in the optimization problem designed for learning the model parameters.
Alternatively, we can come up with sufficient conditions of the structure of the resistive that enforce dissipativity of the construct. We have investigated such conditions in \cite{8814675}, where we showed how the gradients of the energy function should look like. For example, in the case of springs, an energy function of the form $H(x) = \sum_{i=1}^na_ix^{2i}$, with $a_i\geq 0$ it is enough to ensure that the resulting basic construct is dissipative. 
We recall that in the case of the spring, $x$ denotes the spring's elongation. Similarly, we can derive sufficient conditions for the structure of the damper's resistive maps, and mass's energy function. Positive parameter type of constraints (i.e., box constraints) are much easier to deal with in the optimization formulation. Moreover, they can be removed altogether through variable transformations.

\begin{figure}[ht!]
\begin{center}
\includegraphics[width=0.6\textwidth]{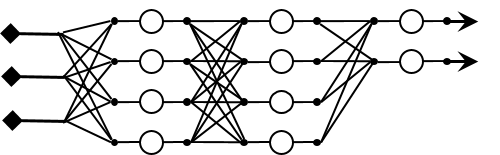}
\end{center}
\caption{Example of a network of p-H constructs with three boundary conditions and two measured quantities.}
\label{fig:01032029}
\end{figure}

We can use the basic p-H construct to build a network of components that can model the observed behavior. An example of such a network is shown in Figure \ref{fig:01032029}, where we depict three boundary conditions, two measured quantities, and three layers. As in the case of a typical NN, choosing the number of layers and the layer sizes, is more of an art than a formal process. We can start with a large enough network and impose sparsity constraints to achieve the simplest network structure that models the observed behavior. We will discuss how we can achieve sparsity in a subsequent section.

The resulting mathematical model is a differential algebraic equation (DAE) that takes the form:
\begin{eqnarray}
\label{eq:01032037}
0 &=& F(\dot{z},z,u;\beta),\\
\label{eq:01032038}
y &=& h(z,u;\beta),
\end{eqnarray}
where $F$ is a map that depends on the state vector $z$ and its time derivative $\dot z$, on the exogenous inputs $u$ through which the boundary conditions are set, and on the vector of parameters comprising the parameters of all energy functions and resistive maps of the basic p-H constructs. The DAE results from collecting the constitutive relations of the basic p-H constructs together with the Dirac structure (energy conservation) constraints.

\section{Training sparse interpretable models}
\label{training_interpretable_models}
It should not be a surprise that learning the parameters of a DAE model is formulated as a (constrained) non-linear least square problem, to which we add regularization function depending on the additional objectives we would like to achieve (e.g., sparse models). The nonlinear least square formulation is given by:
\begin{eqnarray}
\label{eq:01061054}
\min_{w} & J(w) + \lambda R(w)\\
\label{eq:01061055}
\textmd{subject to:} & w\in \mathcal{W},\\
\label{eq:01061056}
& F(\dot{z},z,u;w) = 0, \forall t\in \mathcal{T},\\
\label{eq:01061057}
& y = h(z,u;w), \forall t\in \mathcal{T},
\end{eqnarray}
where $J(w)$ is the quadratic loss function defined in terms of output measurements, e.g., $J(w) = \frac{1}{T}\int_{0}^T\|y_m(t)-{y}(t)\|^2$, where $y(t)$ is the simulated output and $y_m(t)$ is the measured/observed output. The map $R(w)$ is the regularization function (e.g., function that encourages the sparsity of the model or dissipativity of the constructs), $\mathcal{W}$ is a feasibility set for the model parameters, and $\mathcal{T}$ is the time domain over which the output measurements are taken. Note that although the DAE appears as an equality constraint, we actually eliminate it by solving the DAE over the time horizon $\mathcal{T}$, which computes the state vector $z$ and the simulated output measurement vector $y$.

We distinguish two case: (i) the DAE can be transformed into an ordinary differential equation (ODE), and (ii) the DAE cannot be represented as an ODE. The transformation process is based on index reduction methods, which consist of differentiating the algebraic constraints, and then solving the resulting
differentiated system. Because the original constraints have been dropped, and replaced by some of their derivatives, such a process may introduce many additional solutions which no longer satisfy
the constraints. Hence, when possible, modified index reduction techniques based on deflation \cite{oatao9909} are used.

If the dynamical system can be expressed as an ODE, we can use TensorFlow \cite{tensorflow2015-whitepaper} or Pytorch \cite{paszke2017automatic}  ODE solvers capability to compute the latent variables needed for the evaluation of the loss function and its gradients. The advantage of such solvers is that they support automatic differentiation throughout their computation steps.  If the system admits a DAE then we can use DAE solvers that include sensitivity analysis (e.g., IDAS, CVODES). They compute the sensitivity of the latent variables with respect to the optimization variables \cite{10.7717/peerj-cs.54}. Note that the stability of the optimization algorithm depends on the stability of the ODE/DAE solvers. ODE/DAE solvers can become numerically unstable when searching through the parameter space. We can actually learn and avoid regions of the parameter space that cause numerical instability, as we demonstrated in \cite{8431510}. Depending on how the parameter feasibility set $\mathcal{W}$ is modeled there are several options to deal with constraint (\ref{eq:01061055}). For box constraints, we can use variable transformations to eliminate the constraints. Alternatively, we can use projected gradient descent algorithms, provided the projection operation can be done efficiently. If the constraint set is represented through a set of function equalities or inequalities, we can use nonlinear programming (e.g., penalty methods, barrier methods, primal-dual methods, augmented Lagragians methods, etc.) to solve the problem. A particular case of this approach is to include the constraints in the regularization function and test different weights until a satisfactory result is obtained. Unless we are lucky, the result will by suboptimal.

\subsection{Sparsity  and dissipativity constraints}
The sparsity constraints are meant to produce simple models capable of representing the observed behavior or enforce dissipativity for the model elements. Model sparsity can be also achieved after we trained a model, by using model reduction techniques. Sparsity during training can be achieved by cutting the flows on the links. For example, by defining the regularization function as $R(w) = \sum_{j=1}^M|f_j(w)|$, where $f_j$ are the flows through the $M$ links (connections) of the network, we encourage some of the flow to be very small, to the extent that they can be considered zero. We note that the flows $f_j$ are functions of the model parameters and are evaluated by solving the DAE. If we would like to avoid an arbitrary choice for the weight parameter $\lambda$, we can formulate a constraints optimization problem that focuses on reducing the flows through the model, while maintaining a preset accuracy level $\varepsilon$. In particular we have the following constrained optimization problem:
\begin{eqnarray}
\label{eq:01071053}
\min_{w} &  R(w)\\
\label{eq:01071054}
\textmd{subject to:} & J(w)\leq \varepsilon,\\
\label{eq:01071055}
& w\in \mathcal{W},\\
\label{eq:01061056}
& F(\dot{z},z,u;w) = 0, \forall t\in \mathcal{T},\\
\label{eq:01071057}
& y = h(z,u;w), \forall t\in \mathcal{T},
\end{eqnarray}
where $\varepsilon$ can be chosen as a function of the measurement noise variance. This constrained optimization problem can be solved using a primal-dual method, where the primal problem is defined as
\begin{eqnarray}
\label{eq:01071607}
\min_{w} &  R(w)+\lambda_k[J(w)-\varepsilon]\\
\label{eq:01071608}
& w\in \mathcal{W},\\
\label{eq:01071609}
& F(\dot{z},z,u;w) = 0, \forall t\in \mathcal{T},\\
\label{eq:01071610}
& y = h(z,u;w), \forall t\in \mathcal{T},
\end{eqnarray}
and the dual problem is solved by a projected gradient decent algorithm based on the iteration:
$$\lambda_{k+1} = \lambda_k + \alpha [J(w)-\varepsilon],$$
with $\alpha$ the iteration step-size. Note that the primal optimization problem can be transformed into an unconstrained optimization problem by solving the DAE $F(\dot{z},z,u;w)$ at each iteration of the algorithm. If a gradient based optimization algorithm is used, to compute the gradient of the loss function $J(w)$, we need the partial derivatives $\frac{\partial y(t)}{\partial w} = \frac{\partial h}{\partial x}\frac{\partial x}{\partial w}$, along the trajectory of the DAE. We mentioned earlier that in the case the DAE can put represented as an ODE, both TensorFlow and Pytorch can be used. In the case of DAEs, solvers endowed with sensitivity analysis can be used to compute the gradients of the loss function. If the size of the state variables and the number of parameters is large, sensitivity analysis based on backward methods \cite{doi:10.1137/S1064827501380630} (e.g., based on solving the adjoint equations) are recommended.
In the previous section we showed how we can define maps for the basic construct elements that ensure the constructs are dissipative, by imposing a specific structure on the maps and their parameters. Alternatively, we can impose the dissipative condition through a regularization function defined in terms of the construct's instantaneous power. In particular we can define the function $R(w) = \int_{0}^Tp_i(t)dt$, where $p_{i}(t) = e_i^T(t)f_i(t)$ is the instantaneous power of element $i$. 

\section{How we fit in the larger interpretable ML context}
\label{larger_context}
We will use the framework proposed in \cite{Murdoch_2019} to describe the main characteristics of the interpretability approach. Our approach fits the class  of model-based interpretability, where we use constructs that provide insights about the learned relationships. In \cite{Murdoch_2019} the authors state that due to the imposed constraints, the space of possible models is smaller and consequently this may result in lower predictability. Our view is that with the right constructs we can attain high complexity and ensure accurate predictions. As neural networks can approximate arbitrarily close any function by composing layers of linear relations followed by non-linear activation functions, the same can be achieved by composing constructs capable of modeling energy transformations. As in the case of NN-based models we are often interested in finding the simplest  model that is able to accurately describe the observed behavior. For this task we use constraints on the construct flows that effectively eliminate model elements. Here, by \textit{sparsity} we understand the smallest number of constructs that when composed and simulated are able to accurately recover the observed behavior.
Another relevant characteristics is the \textit{simulatability} of the model. 
In our approach, simulatability translates to the ability to simulate the model under different boundary conditions. A sufficient condition that ensure simulatability is stability. Stability follows  by imposing a dissipative constraint on the model constructs \cite{khalil2002nonlinear}. A  component/construct is dissipative if it does not generate energy internally, and all energy is externally supplied. It is sufficient to impose such constraint locally since the composition of dissipative constructs results in a dissipative model.
Our models have \textit{modularity} embedded by default since they are created as compositions of constructs, where each construct can be independently interpreted. For example  energy storage components stores flow or effort.
In addition to the choice of model, the ``features'' used as inputs are also important. In our context the ``features'' are effort or flow sources (e.g., current of voltage sources) that supply energy to the system. They are seen as boundary conditions for the system dynamics. Since they have a particular physical meaning, they are easy to interpret. In the case of autonomous systems (there are no exogenous inputs) there will be no input features but only outputs used in the loss functions. The outputs are functions of  measurements of effort or flow variables.

\section{Examples}
\label{examples}
We demonstrate learning interpretable models for two examples: the inverted pendulum and the Cucker-Smale model for swarm dynamics.

\subsection{Inverted pendulum}
In the first example we show how we can learn the closed-loop behavior of the inverted pendulum using physically interpertable components.

\subsubsection{Physical model}
We are using the inverted pendulum model \cite{matlab-inverted-pendulum}, whose dynamics are given by
\begin{eqnarray}
\label{eqn:06072149}
\dot{x} &=& v\\
\label{eqn:060721450}
\dot{v} &=& \frac{(J+ml^2)(bv-F-ml\omega^2\sin \theta )-m^2l^2g\sin \theta \cos\theta}
{(ml\cos\theta)^2-(m+M)(J+ml^2)}\\
\label{eqn:060721451}
\dot{\theta} &=& \omega\\
\label{eqn:060721452}
\dot{\omega} &=& \frac{ml(F\cos \theta+ml\omega^2\sin\theta\cos\theta-bv\cos\theta+(m+M)g\sin\theta)}{(ml\cos \theta)^2-(m+M)(J+ml^2)}
\end{eqnarray}%
where $x$ and $v$ are the cart's position and velocity, respectively, while $\theta$ and $\omega$ are the pole's angle and angular velocity, respectively. The symbol $F$ denotes the force acting on the cart and plays the role of the input signal. The state vector is given by $z^T=[x,v,\theta,\omega]$. The parameters of the system and their values are listed in Table \ref{tab:09251428}.
 \begin{table}[ht!]
 \caption{Inverted pendulum parameters}
\centering
\begin{tabular}{ |c|c|c|c|c|c|}
 \hline
\textbf{M} & \textbf{m} & \textbf{l} & \textbf{g} & \textbf{J} & \textbf{b}\\
 \hline
 0.5 & 0.2 & 0.3 & 9.81 & 0.006 & 0.1\\
 \hline
\end{tabular}
\label{tab:09251428}
\end{table}
Note that in its original form, the dynamics of the inverted pendulum system are represented as a DAE. We explicitly solved for the accelerations to generate the ODE form.

The inverted pendulum system has two inertial elements (the cart with mass $M$ and the pendulum with mass $m$) and at least one resistive element due to the cart friction. The translational motion of the cart is transform into a rotational motion that acts upon the pendulum. Hence our model will have two inertial elements (a translational one and a rotational one), a resistive element and a transformer.  The  inertial elements store kinetic energy, while the resistive element dissipates energy. We recall that the effort and flow variables for the translational and rotational cases are forces ($f$) and velocities ($v$), and torques ($\tau$) and angular velocities ($\omega$) respectively. We recall that the translational and rotational elements are given by
$$\begin{array}{ccc}
\dot{\xi} &=& f,\\
v &=& \frac{\partial H}{\partial \xi},
\end{array}
\begin{array}{ccc}
\dot{\eta} &=& \tau,\\
\omega &=& \frac{\partial H}{\partial \eta},
\end{array}
$$%
respectively. We will consider resistive elements for both the translational and rotational elements, namely $f = R_t(v)$ and $\tau = R_r(\omega)$. Finally the transformer element can be described by a map $g:\mathds{R}^4\rightarrow \mathds{R}^2$, such that $g(f,v,\tau,\omega) = 0$.

\subsubsection{Model training}
\begin{figure}[ht!]
        \centering
        \includegraphics[width=0.6\textwidth]{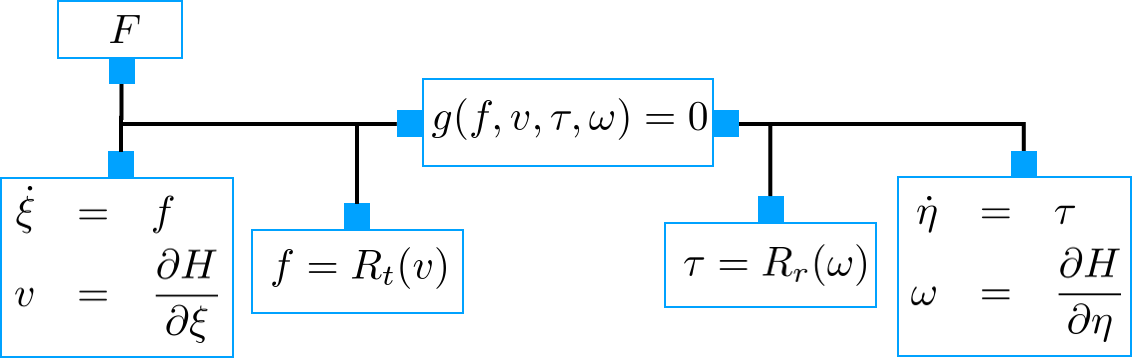}
        \caption{Block diagram for the inverted pendulum surrogate model in p-H formalism.}
        \label{fig:p-H diagram}
\end{figure}

By combining the five types of p-H elements we obtain the configuration shown in Figure \ref{fig:p-H diagram}, where we added a force source that generates the force $F$ acting on the cart.
Our objective is to learn the component maps, where we make the common sense assumption that the inertial and resistive elements are linear, that is: $v = \frac{\xi}{M}$, $\omega = \frac{\eta}{J}$, $f = d_t v$ and $\tau = d_r \omega$, respectively. The key component is the transformer element. If you model the map $g$ as a nonlinear map, we end up with a DAE which will prevent us from using ODE solvers endowed with automatic differentiation feature. When solving the DAE, the map $g$ corresponds to an algebraic loop for which we need to use a Newton-Rhapson algorithm to compute the variables $f$ and $\tau$. Note that the velocities are states and hence considered known for the purpose of solving the DAE at each time instant. The solution of the algebraic loop is a map $(f,\tau) = h(v,\omega)$, 
Hence, we can replace the map $g$ with two maps $f = h_1(v,\omega)$ and $\tau = h_2(v,\omega)$. This change transforms the DAE into an ODE. In a sense, instead of learning the map $g$, we learn the solution of map $g$ along the system trajectory, where the unknown variables are the force $f$ and torque $\tau$. The resulting ODE model is given by

\begin{eqnarray*}
\dot{x} &=& v\\
\dot{v} &=& \frac{1}{M}(-F-d_1v+f)\\
\dot{\theta} &=& \omega\\
\dot{\omega} &=& \frac{1}{J}(\tau -d_2\omega),
\end{eqnarray*}%
where $f = h_1(v,\omega)$ and $\tau = h_2(v,\omega)$.
We parameterize the map $h = h(v,\omega;\beta)$ as a neural network with one hidden layer of size $50$, where the activation function is chosen as {\tt tanh}. Hence, the total number of parameters is $(50\times 2+50 + 2\times 50 +2)+4=156$ where the last $4$ parameters correspond to the linear inertial and resistive components. We use Pytorch \cite{paszke2017automatic} and the {\tt torchdiffeq} Python package \cite{chen2018neural} to solve a nonlinear least square optimization problem. We assume that the state vector is measured and  that the input force  $F$ is generated using a pre-trained linear controller with gain $K=[1.2501,   2.7612, -16.3099,  -3.7814]$.
Based on the diagram shown in Figure \ref{fig:p-H diagram} we constructed an ODE and trained the components parameters  using four time series with initial conditions belonging to $\theta_0\in \{-0.1, 0.1\}$, $x_0\in \{-0.2, 0.2\}$ and zero velocities. In particular, we solved the following optimization problem
\begin{eqnarray*}
\min_{\beta} & & \frac{1}{N}\sum_{i=1}^N\sum_{j=1}^T\|z^{(i)}(t_j)-\hat{z}^{(i)}(t_j)\|^2  + \lambda \|ODE(0)\|^2\\
\textmd{subject to:} & & \dot{\hat{z}} = ODE(\hat{z};\beta), \ \hat{z}(0)=z(0),
\end{eqnarray*}
where $ODE(z)$ is the ODE generated by the block diagram shown in Figure \ref{fig:p-H diagram}. The loss function includes a regularization function that ensured the model has an equilibrium point at zero.

\begin{figure}[ht!]
        \centering
        \includegraphics[width=0.7\textwidth]{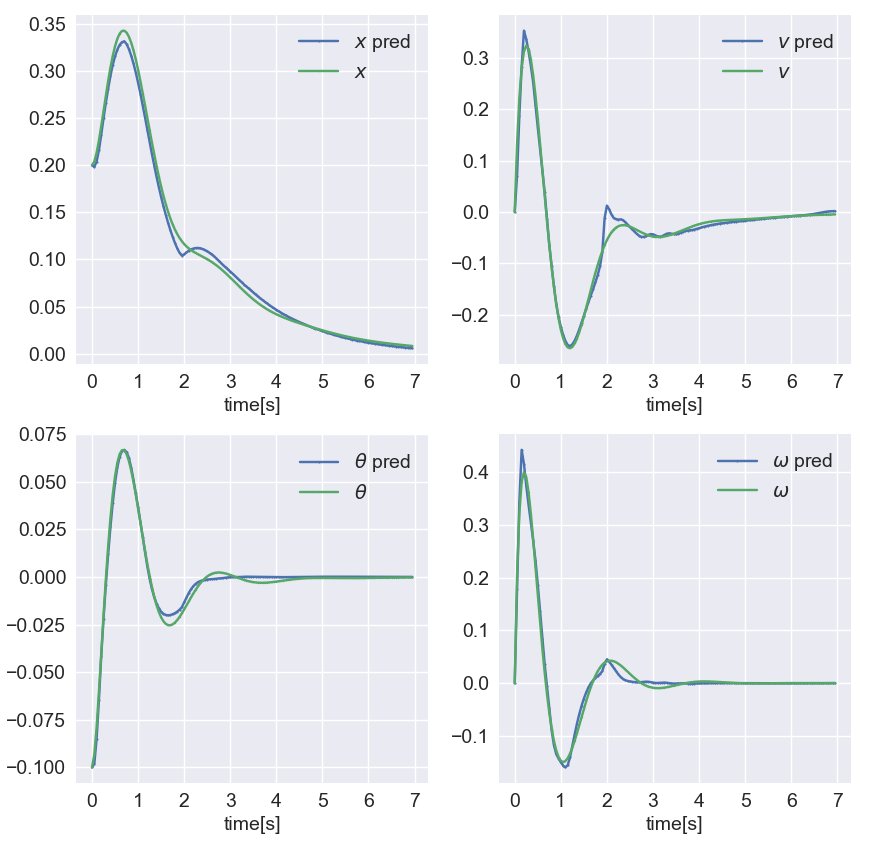}
        \caption{Inverted pendulum trajectories: physics-based model (\ref{eqn:06072149})-(\ref{eqn:060721452}) (green) versus trained model using p-H components (blue). Training MSE=$8.45\times 10^{-5}$.}
        \label{inverted_pendulum_pH}
\end{figure}

We used  Adam \cite{Adam} algorithm to learn the parameters of the p-H components. In this type of learning, the time complexity corresponding to one optimization iteration includes the time for solving the ODE (forward propagation) and the time for computing the sensitivities of the state vector with respect to the optimization parameters. The latter is typically larger. In addition, the choice of ODE solver and the solution sampling period makes a significant difference. We chose an ODE solver based on the {\tt midpoint} method. Using a time horizon of 6 seconds and a sampling period of 0.05 seconds, the time for each optimization iteration is $1.2$ seconds, where more than 80\% is allocated for computing the state sensitivities. We have experimented with other ODE solvers but they typically have higher time complexity. For example, when using the fourth order Runge-Kutta method, the iteration time is more than 2.3 seconds. The iteration time increases with the length of the time horizon and the sampling frequency. A comparison between a simulated (``true'') and learned system trajectory is shown in Figure \ref{inverted_pendulum_pH}.
We tested the generalizability to one hundred different initial conditions randomly drawn so that $x_0\in [-0.2,0,2]$, $v_0\in [-0.1,0.1]$, $\theta_0\in [-0.2,0.2]$ and $\omega_0\in [-0.1,0.1]$. We computed the MSEs for each initial condition. The statistics of the resulting MSE values is:  the mean is given by $1.57\times 10^{-4}$, and the standard deviation is given by $1.4\times 10^{-3}$. Note that the key aspect of learning this model is the velocity trajectories which are the inputs for the map $h$. The more diverse velocity trajectories we use, the more generalizable the model will be.

\subsection{Swarm dynamics}
We demonstrate how the p-H formalism can be used to learn the interaction behavior between particles.
\subsubsection{Mathematical model}
\begin{figure}[ht!]
        \centering
        \includegraphics[width=0.45\textwidth]{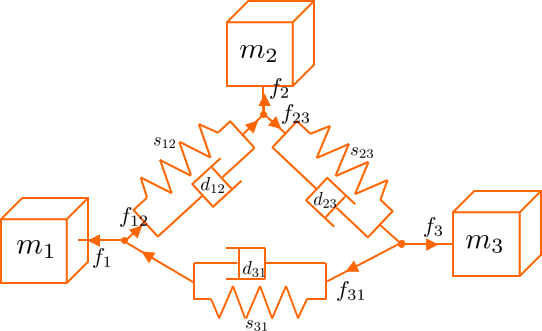}
        \caption{Fully connected, 3-dimensional mass-spring-damper network}
        \label{fig:MSD}
\end{figure}

We are interested in models describing the dynamics of swarms or \textit{particle} ensembles (e.g. bird flocks),
which have been studied intensively through the years  \cite{carrillo_ParticleKineticHydrodynamicModelsOfSwarming_2010,carrillo_newInteractionPotentialForSwarms_2013,reynolds_Boids_1987}.
We model the system of interacting particles as
a graph topology based on port-Hamiltonian basic constructs, and learn the parameters of the constructs. To showcase our approach we use the Cucker-Smale (CS) model
\cite{carrillo_ParticleKineticHydrodynamicModelsOfSwarming_2010} to generate training and test data for the learning tasks.
Let $i$ denote a particle in an ensemble of $N$ particles. The CS particle interaction model that include particle interactions based on potential energy as well
\cite{carrillo_ParticleKineticHydrodynamicModelsOfSwarming_2010} is given by
\begin{eqnarray}
  \label{eq:22020224}
  \dot{x}_i& =& v_i\\
     \label{eq:22030224}
  \dot{v}_i& =& \frac{1}{N}\sum_{j=1}^NG(\|x_i-x_j\|)(v_j-v_i)-\frac{1}{N}\sum_{i\neq j}\nabla U(\|x_i-x_j\|)
\end{eqnarray}
where a typical choice for the interaction function $G$ is $G(r) = \frac{1}{(1+r^2)^{\gamma}}$, and the potential function takes the form $U(r) = -C_Ae^{-r/l_A}+C_Re^{-r/l_R}$, with $C_A,C_R,l_A,l_R$ positive scalars. Note that in the case of the CS model it makes sense to use the p-H constructs. Indeed, consider the three particle example for the one dimensional case, where the grounded dampers and spring are omitted. This can be achieved by considering linear spring-dampers with zero coefficient. The fully connected topology of the mass-spring-damper network is shown in Figure \ref{fig:MSD}.
We denote by $H_i$ and $H_{ij}$ the Hamiltonian functions of the masses and springs, respectively. We assume unitary masses, and hence the momenta are equal to the mass velocities, that is, $p_i = v_i$, $i=\{1,2,3\}$. The forces through the links are the sum of the forces through the dampers and springs, and are given by $f_{ij} = \frac{\partial H_{ij}}{\partial q_{ij}} + R(q_{ij})(v_i-v_j)$,
for $(i,j)\in\{(1,2),(2,3),(3,1)\}$. The forces through the masses can be expressed as: $f_1 = f_{31}-f_{12}$, $f_2 = f_{12}-f_{23}$ and $f_3 = f_{23}-f_{31}$.  We get the expressions for the mass momenta dynamics as:
\begin{eqnarray}
\label{eq:13110221}
\dot{p}_1 = \frac{\partial H_{31}}{\partial q_{31}}-\frac{\partial H_{12}}{\partial q_{12}}+ R(q_{31})(v_3-v_1)+R(q_{12})(v_2-v_1),\\
\label{eq:13120222}
\dot{p}_2 = \frac{\partial H_{12}}{\partial q_{12}}-\frac{\partial H_{23}}{\partial q_{23}}+ R(q_{12})(v_1-v_2)+R(q_{23})(v_3-v_2),\\
\label{eq:13110223}
\dot{p}_3 = \frac{\partial H_{23}}{\partial q_{23}}-\frac{\partial H_{31}}{\partial q_{31}}+ R(q_{23})(v_2-v_3)+R(q_{31})(v_1-v_3).
\end{eqnarray}
The dynamics for the spring elongations are
\begin{equation}
\label{eq:13070222}
\dot{q}_{ij} = v_i-v_j = \frac{\partial H_i}{\partial p_i}-\frac{\partial H_j}{\partial p_j}
\end{equation}
for $(i,j)\in\{(1,2),(2,3),(3,1)\}$.
To recover the CS model with potential, we replace the relative positions $q_{ij}$ with the absolute positions, namely $q_{ij} = q_{i}-q_{j}$. Recalling that spring potentials are symmetric functions, we get that
{\small
\begin{eqnarray}
\label{eq:14110222}
\frac{\partial H_{31}}{\partial q_{31}}-\frac{\partial H_{12}}{\partial q_{12}} = -\frac{1}{3}\left(\nabla U(q_1-q_3)-\nabla U(q_1-q_2)\right)\\
\label{eq:14110223}
\frac{\partial H_{12}}{\partial q_{12}}-\frac{\partial H_{23}}{\partial q_{23}} = -\frac{1}{3}\left(\nabla U(q_2-q_1)-\nabla U(q_2-q_3)\right)\\
\label{eq:14110224}
\frac{\partial H_{23}}{\partial q_{23}}-\frac{\partial H_{31}}{\partial q_{31}} = -\frac{1}{3}\left(\nabla U(q_3-q_2)-\nabla U(q_3-q_2)\right)
\end{eqnarray}}
Substituting (\ref{eq:14110222})-(\ref{eq:14110224}) in (\ref{eq:13110221})-(\ref{eq:13110223}), and recalling that under our assumptions $p_i=v_i$, we recover  exactly the CS model with potential. Hence we showed that the CS dynamics can be modeled and trained using constructs from the p-H formalism. In particular, we can explain the swarm dynamics: each particle behaves like a flow store, and the interaction between particles can be understood as a combination of effort storage (the spring) and energy dissipation (damper).  Tn the case of homogeneous particles, the training process is simplified since all energy functions and resistive maps have identical parameterizations.

\subsubsection{P-H formalism based model}
We assume that we measure the particle trajectories and the objective is to learn and interpret how the particle interact. We model the interaction between the particles as a combination of resistive and potential maps. The ``force'' that controls how particle interact is modeled as a combination of resistive and potential interaction. We consider the force expression as $F(p,q) = \frac{\partial H}{\partial q}(p,q)+R(p,q)$, where $p$ and $q$ are the relative momentum and distance, respectively. Since we cannot measure the resistive and potential effects separately, we model the overall effect of the two phenomena. In particular, we represent the force $F$ as $F(p,q) = f(p,q;\beta)$, where $f$ is a function of $p$ and $q$, and depends on a vector of parameters $\beta$. It follows that we have the following model:
\begin{eqnarray}
    \label{eq:02340224}
    \dot{q}_i &=& p_i\\
    \label{eq:22350224}
    \dot{p}_i &=& \frac{1}{N}\sum_{j=1}^N F(p_j-p_i,q_j-q_i;\beta),
\end{eqnarray}
where $p_i$ and $q_i$ are the particle momentum and position, respectively. We  assume unitary mass, and hence the momentum can be interpreted as the particle velocity. In the original CS model, the potential function governing the behavior of the nonlinear springs depend on the relative position only. In addition, the resistive map describing the damping effect is zero for zero relative velocity. Hence, we can recover the force generate by the nonlinear force by evaluating the force $F(p,q)$ at $p=0$, that is, $F(0,q)$.
We generated is represented by trajectories generated by the CS model. We consider 100 particles in the model and assume we can measure the positions and velocities of the particles. The parameters for the CS model were chosen as: $\gamma = 0.15$, $C_A=200$, $l_A=100$, $C_R=500$, $l_R=2.0$. Figure \ref{fig:CS_trajectories} shows the velocities of the first 10 particles as generated by the CS model.
\begin{figure}[ht!]
        \centering
        \includegraphics[width=0.8\textwidth]{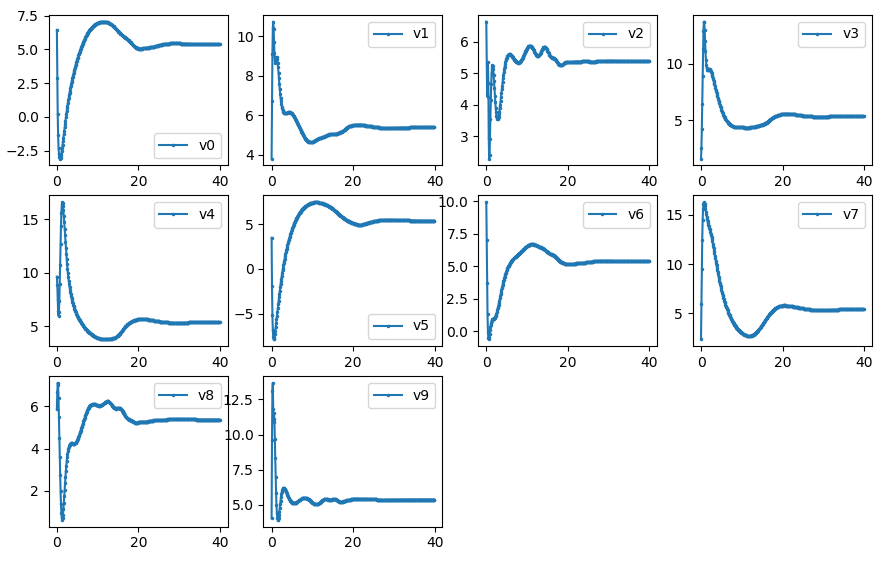}
        \caption{CS model generate trajectories: $x$-axis represents the time in seconds, and $y$-axis represents the velocity in m/sec.}
        \label{fig:CS_trajectories}
\end{figure}
\subsubsection{P-H formalism based model training}
\begin{figure}[ht!]
        \centering
        \includegraphics[width=0.6\textwidth]{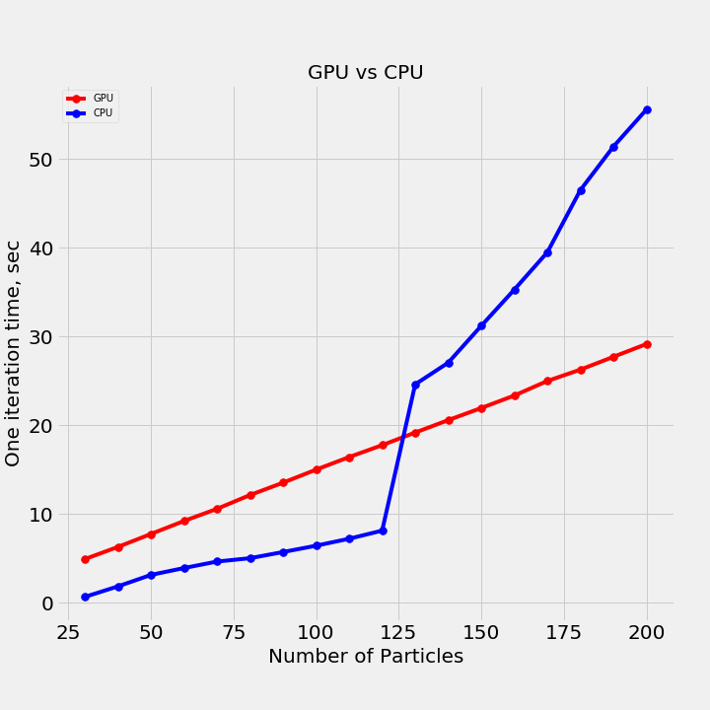}
        \caption{Time per optimization iteration when learning the interaction function: CPU vs. GPU }
        \label{fig:complexity}
\end{figure}

\begin{figure}[ht!]
        \centering
        \includegraphics[width=0.8\textwidth]{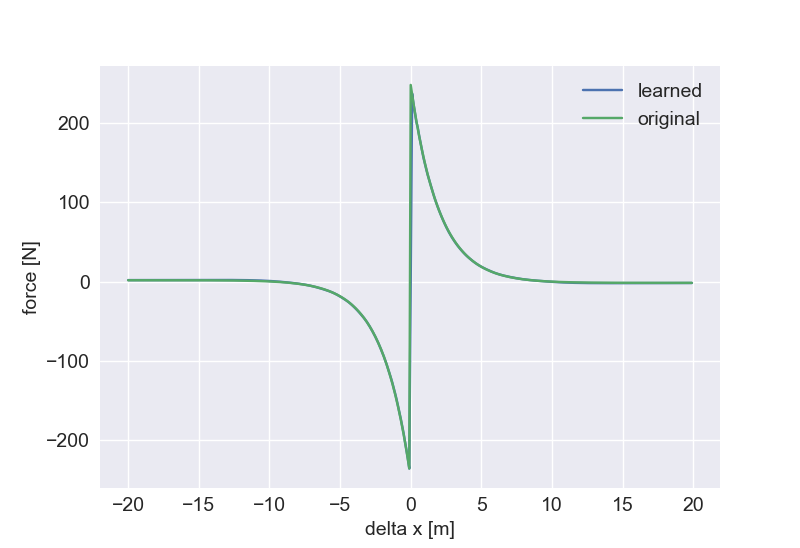}
        \caption{Comparison between the original CS (green) and learned (blue) potential functions (MSE=0.36).}
        \label{fig:potential functions}
\end{figure}

To learn the force map $F(p,q)$ we minimize solve the following optimization function
\begin{eqnarray}
\label{equ:02061419}
\min_{\beta} & &\frac{1}{n}\sum_{l,i}^n \|{p}_i^{(l)}-\hat{{p}}_i^{(l)}\|^2 + \|{q}_i^{(l)}-\hat{{q}}_i^{(l)}\|^2 + \lambda \|F(0,0;\beta)\|^2\\
\label{equ:02061420}
\textmd{subject to:} & & \dot{\hat{q}}_i = \hat{p}_i,\\
\label{equ:02061421}
& & \dot{\hat{p}}_i = \frac{1}{N}\sum_{j=1}^N F(\hat{p}_j-\hat{p}_i,\hat{q}_j-\hat{q}_i;\beta),
\end{eqnarray}
where $p_i$ and $q_i$ are the measured particle momenta and positions, and $\beta$ is the vector of parameters of the force map $F(p,q)$. In addition to the quadratic loss function, we added a regularization function that enforces the zero behavior of the interaction function.
As in the case of the inverted pendulum, the time complexity comes from the solving the ODE governing the particle interaction rather than from the number of optimization parameters. We modeled the interaction function by a one hidden layer neural network. The size of the hidden layer is 100 and we use {\tt tanh} as the nonlinear activation function.
We used Pytorch to train the model parameters and {\tt torchdiffeq} to solve the ODE corresponding to the CS model. We experimented with different number of particles and ODE solvers. Figure \ref{fig:complexity} shows the time complexity per optimization iteration as a function of number of particles, when using CPUs and GPUs, and {\tt dopri5} as ODE solvers.
Note that although the time complexity in the GPU case increases linearly, we still have large numbers for iteration. After experimenting with other ODE solvers, we chose one based on the {\tt midpoint} method since it provides a good trade off between complexity and accuracy. The reduced complexity if beneficial in particular for the back propagation step.
We generated 10 time series using the CS-model with 30 particles and trained the parameters of the interaction model using a stochastic version of the Adam algorithm, where at each iteration we chose one of the 30 time series at random.  The time horizon of the time series is 40 second and the sampling period is 0.1 seconds. The initial conditions for the particle positions and velocities were chosen at random in the interval [-10,10].
Figure \ref{fig:CS_trajectories} shows a comparison between the velocity trajectories generated by the CS-model and the trajectories generated by the CS-model with learned interaction functions. The initial conditions were chosen at random. The two sets of trajectories match both qualitatively and quantitatively.
\begin{figure}[ht!]
        \centering
        \includegraphics[width=0.9\textwidth]{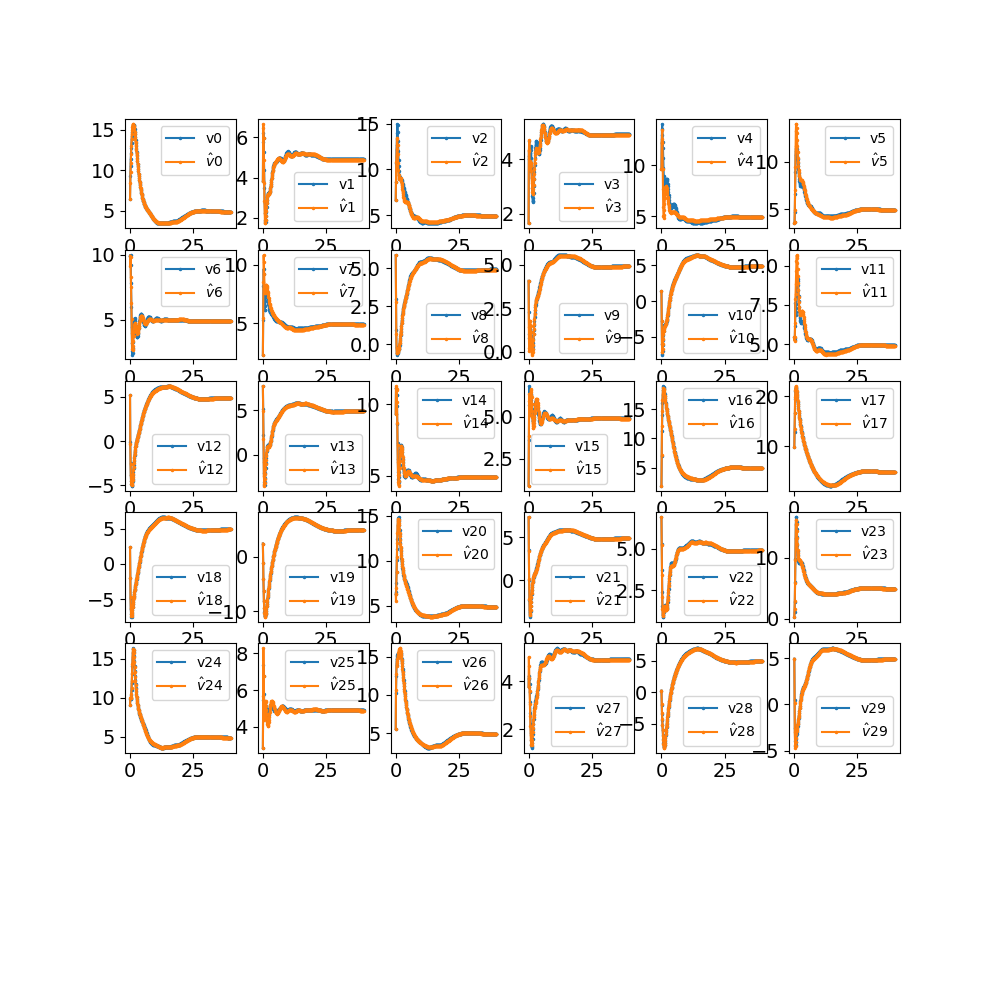}
        \caption{Simulated (blue) vs. predicted (orange) particle velocity trajectories (MSE = 0.03147).}
        \label{fig:CS_trajectories}
\end{figure}
We executed fifty more experiments for validation purposses, where the initial conditions where chose at random. The MSE statistics are: the mean is 0.06 and the standard deviation 0.04. The numbers do not appear to be very small. We have to recall though that we only used 10 time series and that the generalizability of the interaction function depends on what relative positions and velocities are hit during the particle evolution.
We executed another validation experiment, meant at checking if we can recover the force component generated by the potential energy. We evaluated the interaction function by varying the relative position while setting the relative velocity to zero. We compared the learned potential function with the potential function  defined by the CS model. A graphic comparison is shown in Figure \ref{fig:potential functions}. Except around the origin, the two function are almost identical, and demonstrate that indeed we learn a repulsive behavior when particle get too close to each others.

\subsection{Time complexity analysis}
\label{time complexity}
We distinguish between two situations: the ODE and the DAE cases. In the ODE case, and when using automatic differentiation, the time complexity is split in two parts: the state vector computation  and evaluation of the state sensitivity with respect to the optimization parameters. Assume that loss function $\mathcal{L}$  depends on the state vector $x_k$ and optimization parameters $\beta$ and that the discrete representation of the state vector is given by $x_{k+1} = f(x_{k};\beta)$. Then the derivative of the loss function with respect to the optimization parameters is $\frac{d \mathcal{L}}{d\beta} = \sum_{k=0}^T\frac{\partial \mathcal{L} }{\partial x}(x_k;\beta)\frac{d x_k}{d \beta}$, where we assume that the loss function does not explicitly depends on $\beta$. To evaluate the gradient of $\mathcal{L}$ we require the state and the state sensitivity  evolution for $T$ time steps. The state evolution is determined using the iteration $x_{k+1} = f(x_{k};\beta)$ that incurs $O(c_1\times T)$ time complexity, where $c_1=c_1(n)$ is the time required to evaluate the map $f$ at $x_k$ and $\beta$, and depends on the state vector size equal to $n$. The sensitivities evolve according to the dynamics $\frac{d x_k+1}{d \beta} = \frac{\partial f}{\partial x}(x_k;\beta)\frac{d x_k}{d \beta}+\frac{df}{d\beta}$. Hence, the time necessary to evaluate the sensitivities is $O(c_2\times T)$, where $c_2 = c_2(n\times m)$ is the time necessary to evaluate the current sensitivity, where $m$ is the size of vector $\beta$. Note that $c_2$ is a function of $n\times m$ since $\frac{d x_k}{d \beta}$ is a matrix. Hence, then time per optimization iteration is $O((c_1+c_2)\times T)$, while the total optimization time complexity is $O((c_1+c_2)\times T\times N)$, where $N$ is the number of iterations. There are three  parameters that determine the total time complexity: the number of optimization parameters ($m$), the number of states ($n$) and the state trajectory length ($T$). Note that both $c_1$ and $c_2$ depend also on the type of ODE solvers, since some solvers employ more complex discretization schemes.

In the case the dynamics is represented as a DAE, the solvers need to invert the system Jacobian at every time iteration, an operation that takes $O(n^3)$. The Jacobian serves for computing the sensitivities as well \cite{PETZOLD20061553}. The inverted Jacobian is further used in a Newton-Rhapson algorithm that solves a nonlinear system of equations. In this case, the time complexity for the optimization algorithm is $O((c_1+c_2)\times T\times N)$, where $c_1$ is $O(n^3)$ and $c_2$  is  $O(n\times m)$. We note that the size of the state vector has a more significant impact on the time complexity. There are several optimization steps that can be performed to the Jacobian to reduce the size of the matrix that needs to be inverted. In particular, after the system of equations is brought to the block lower triangular form (BLT), the blocks on the diagonal are further simplified through a ``tearing'' operation that is based on Pantelides's algorithm. These steps are case by case, and we cannot guarantee that they can always result in a reduction of the matrix size that needs to be inverted. A particular case is when the  loss function depends on output measurements and not explicitly on the state vector.  In this situation we can use backward methods \cite{doi:10.1137/S1064827501380630,10.1115/DETC2005-85597} to compute the sensitivity of the output measurements with respect to $\beta$, without having to explicitly compute the state vector sensitivities. Still the Jacobian inversion cannot be escaped, and therefore for large values of $n$ and $T$ it becomes the dominant component of the time complexity.

\section{Conclusions}
We showed how we can use the p-H formalism to learn physically interpretable system models. Basic p-H elements serve as blueprints to construct complex model architectures able to reproduce behaviors of physical systems. We discussed properties of models that ensure numerical stability and introduced algorithms and custom regularization functions that encourage model sparsity. We showcased our approach on two examples: the inverted pendulum and swarm dynamics based on the C-S model. We evaluated the time complexity of the learning algorithm. In the case the system dynamics is expressed as a DAE that cannot be reduced to an ODE, the learning procedure is numerically costly, especially for large number of states. We believe that to ensure the scalability of the learning process, we need to come up with efficient algorithms and hardware architectures that can keep the complexity in check.

\bibliographystyle{plain}
\bibliography{references}

\end{document}